\documentclass{article}
\usepackage{amsmath}
\usepackage{amsfonts}
\usepackage{mathtools}
\usepackage{float}
\usepackage{subfig}

\usepackage{graphics} 
\usepackage{times}    
\usepackage{url}      
\usepackage{authblk}

\makeatletter
\def\url@leostyle{%
  \@ifundefined{selectfont}{\def\UrlFont{\sf}}{\def\UrlFont{\small\bf\ttfamily}}}
\makeatother
\DeclareMathOperator*{\argmin}{arg\,min}
\DeclareMathOperator*{\argmax}{arg\,max}

\def\pprw{8.5in}
\def\pprh{11in}

\usepackage[]{hyperref}
\hypersetup{
pdftitle={SIGCHI Conference Proceedings Format},
pdfauthor={LaTeX},
pdfkeywords={SIGCHI, proceedings, archival format},
bookmarksnumbered,
pdfstartview={FitH},
colorlinks,
citecolor=black,
filecolor=black,
linkcolor=black,
urlcolor=black,
breaklinks=true,
}

\begin{document}

\title{Effective sampling for large-scale automated writing evaluation systems}

\author[1]{Nicholas Dronen\thanks{dronen@colorado.edu}}
\author[2]{Peter W. Foltz\thanks{peter.foltz@pearson.com}}
\author[3]{Kyle Habermehl\thanks{kyle.habermehl@pearson.com}}
\affil[1]{University of Colorado \& Pearson \\ Boulder, CO, USA 80301}
\affil[2]{Pearson \& University of Colorado \\ Boulder, CO, USA 80301}
\affil[3]{Pearson \\ Boulder, CO, USA 80301}

\renewcommand\Authands{and}

\maketitle

\begin{abstract}
Automated writing evaluation (AWE) has been shown to be an effective
mechanism for quickly providing feedback to students.  It has already
seen wide adoption in enterprise-scale applications and is starting
to be adopted in large-scale contexts.  Training an AWE model has
historically required a single batch of several hundred writing
examples and human scores for each of them.  This requirement limits
large-scale adoption of AWE since human-scoring essays is costly.
Here we evaluate algorithms for ensuring that AWE models are
consistently trained using the most informative essays.  Our results
show how to minimize training set sizes while maximizing
predictive performance, thereby reducing cost without unduly
sacrificing accuracy.  We conclude with a discussion of how to
integrate this approach into large-scale AWE systems.
\end{abstract}

\section{Introduction}
Automated writing evalution (AWE) is the use of natural language
processing and statistical modeling to score samples of writing.
The first model for automatically scoring essays was created on
punch cards in the 1960s \cite{Page1968-ix}.  At present there are
several commercial automated essay scoring operations \cite{Rudner2006-qb,
Landauer2003-vm, Attali2006-nz}.  They are often used as the sole
scorer in formative applications \cite{WriteToLearn,ETSCriterion}
and increasingly in summative assessments (e.g. Pearson Test of
English Academic).  Even wider adoption of automated writing
evaluation appears to be underway.  Recently, for example, the
Hewlettt Foundation sponsored open competitions for essay-length
and short answer AWE modeling on Kaggle.com to help foster greater
adoption of the technology \cite{KaggleAES, KaggleSAS}.  The edX
platform for massive open online classes (MOOCs) contains a system
for automated writing evaluation \cite{edXorg, edXdiscern, edXease}.

Typically, an AWE model is trained using supervised learning and
performs best when it has been trained for a specific prompt and
scoring trait \cite{Foltz2013-li}.  Training an AWE model starts
with a batch of essays and one or more human scores for each essay.
In the usual training procedure, the essays are preprocessed by a
natural language processing pipeline, which results in a feature
vector for each essay.  The feature vectors and scores are then
input to a learning algorithm that learns a model mapping feature
vectors to scores.  When the feature vector of a previously unseen
essay is presented to the model, it can immediately provide a score.

Historically, the process of training an AWE model has required at
least several hundred human-scored essays.  The purpose of this
requirement is twofold: to reduce sampling noise in the training
set and to ensure that the model performs about as well as one
trained with many more essays.  In an online setting, however, a
system for collecting essays and human scores can cause an AWE model
to be trained whenever a new human score is entered into the system.
For example, the edX AWE system allows models to be trained with
as few as ten essays \cite{edXdiscerngithub}.  If the training set
is sampled randomly, then training with so few essays will -- with
high probability -- yield an AWE model that performs poorly.

The challenge of adopting AWE in large-scale contexts is that the
best way to use the technology -- with a customized scoring model
for each prompt -- is also quite expensive.  In an enterprise
environment, the cost of scoring ranges from \$3-6 USD per essay.
Assuming 500 essays for a training set and 250 for test, the cost
per prompt is \$2250-4500.  In this paper we show how to overcome
this key barrier to the widespread adoption of AWE in large-scale
contexts.  The method we use minimizes the number of essays that
need to be scored and maximizes the information that each training
example provides.  This can also enable an online data collection
system to intelligently choose which essays to be scored by humans.
We conclude by discussing how to effectively integrate this method
into AWE systems.

There are two bodies of literature about effective methods for
choosing samples for training a supervised model.  The older is
\textit{optimal experimental design} \cite{Fedorov1972-pg}, a subfield of
statistics that originated with the work of Kirstine Smith in the
early 20th century \cite{Smith1918-ho}.  The paradigmatic supervised
learning algorithm for this literature is linear regression.  In
machine learning, \textit{active learning} occupies much the same
space as optimal design, although the literature tends to focus 
more on classification than regression \cite{Settles2010-wp}.

The usual scenario for optimal experimental design is when samples
are expensive to obtain, as with clinical trials or gathering samples
during field studies in remote areas.  An optimal design algorithm
chooses samples from a pre-specified set of feature vectors in such
a way as to minimize the variance in the supervised model.  In
active learning the assumption is usually that one begins
with a set of \textit{unlabeled} samples and wishes to obtain labels
for them.  For example, unlabeled samples can be HTML documents
obtained by crawling the web and the desired labels can be the
topics of the documents.  Papers about regression in the active
learning literature often cite the optimal design work of Valerii
Fedorov (e.g.  \cite{MacKay1992-gp,Sugiyama2005-jr,Bach2007-ub}).

When using an algorithm to choose samples prior to training a
supervised learning algorithm, obtaining good results requires that
the assumptions of the sampling algorithm match the assumptions of
the learning algorithm.  Here we use parametric linear regression
and we employ an algorithm from the optimal experimental design
literature \cite{Fedorov1972-pg} that is derived from the same
foundations as linear regression.

Since we start with essays and ask which ones should be graded,
ours is really the classic active learning scenario.  We, however,
stick mostly to the terminology from the optimal design literature
throughout this paper.

\section{Data sets}

An AWE data set consists of some number of written responses and a
set of scores for each response.
In our experiments we use the industry track data
from the Automated Student Assessment Prize (ASAP) Automated
Essay Scoring (AES) competition sponsored by the Hewlett Foundation
held in 2012 \cite{KaggleAES}.  It contains training and test sets
numbered 1 through 8.  Key parts of the rubric of each set are summarized
in Table \ref{tab:asapaes-rubric}.   

In sets 1 and 2, a student is asked to write persuasively.  Essay
set 1 has the student take a position on the effects of computers
on society and write a letter to the editor of a newspaper arguing
for this position.  Sets 3-6 are \textit{response to source} prompts;
they have the student read and analyze some source text and provide
evidence from the text to support their analysis.  The source of
essay set 5 is a passage about the son of Cuban immigrants growing
up in New Jersey.  The student must describe the mood of the text.
The narrative task tends to be more open ended.  In essay set 8,
the task is to write about an experience in which laughter was an
important element.

The target variable of the data sets is the score or sum of scores
that an essay was given by one or more human raters.  The range of
the target variable for each essay set is shown in Table
\ref{tab:asapaes-rubric}.  Essay set 2 has two target variables,
which we identify as $2a$ and $2b$ depending on which target variable
the model was trained to predict.  For all essay sets the target
variable includes all integers between the minimum and maximum
scores.  The target variable for essay set 7, for example, is the
integers 2, 3, \dots, 30.

The size of the training and test sets are shown in Table
\ref{tab:asapaes-data}.  While relatively small, these test sets
are nonetheless sufficient to ensure that the performance
we report in the results section approaches the performance one
might expect on a much larger sample.  The same table shows the
rounded average number of words in each essay set.  A peculiarity
of these sets is that in some cases the average length is quite 
low.  This is true for all of the response-to-source sets, suggesting
that students write less when performing a more focused task.

Before using an algorithm to select which essays
raters should score, we compute the feature vector for each essay.
In our experiments we use a subset of twenty-eight features --
related to mechanics, grammar, lexical sophistication, and style
-- from the Intelligent Essay Assessor \cite{Foltz2013-li}.

\begin{table}[!htbp]
\caption{ASAP AES rubric}
\label{tab:asapaes-rubric}
\centering
\begin{tabular}{r|crr}
  \hline
  Set & Task & Grade & Range \\
  \hline
  1  & Persuasive &  8 & 2-12 \\ 
  2a & Persuasive & 10 & 1-6 \\ 
  2b & Persuasive & 10 & 1-4 \\ 
  3  & Source     & 10 & 0-3 \\ 
  4  & Source     & 10 & 0-3 \\ 
  5  & Source     &  8 & 0-4 \\ 
  6  & Source     & 10 & 0-4 \\ 
  7  & Narrative  &  7 & 2-30 \\ 
  8  & Narrative  & 10 & 0-60 \\ 
   \hline
\end{tabular}
\end{table}

\begin{table}[!htbp]
\caption{ASAP AES data summary}
\label{tab:asapaes-data}
\centering
\begin{tabular}{r|rrr}
  \hline
  Set & Train $n$ & Test $n$ & Avg. Words \\
  \hline
  1   & 1785 & 589 & 372 \\ 
  2a  & 1799 & 600 & 386 \\ 
  2b  & 1799 & 600 & 386 \\ 
  3   & 1699 & 568 & 110 \\ 
  4   & 1763 & 586 & 96 \\ 
  5   & 1805 & 601 & 124 \\ 
  6   & 1800 & 600 & 154 \\ 
  7   & 1469 & 495 & 174 \\ 
  8   &  917 & 304 & 636 \\ 
   \hline
\end{tabular}
\end{table}

\section{Regression modeling}

The target variable for automated essay scoring tasks is the student's
score on an essay.  It is typically an integer in a relatively
narrow range determined by the scoring rubric.  The ordering inherent
in the target variable implies that essay scoring data are best
modeled using a regression algorithm.  Here we describe the approach
we take to regression modeling in our experiments.

When the length of feature vectors in a design matrix $\xi$ exceeds
the number of feature vectors, the ordinary least squares solution
is ill-posed.  In this paper we allow $m$, the number of feature
vectors to be as small as ten.  Since $p$, the number of features,
is twenty-eight, the ordinary least squares solution would be
suboptimal for some of the models we train.  Thus we use regularized
regression instead of ordinary least squares.  The regularized
linear regression solution for $\beta$ is the solution for ordinary
least squares augmented with a penalty function $J(\beta; \lambda)$,
\begin{equation*}
\hat{\beta}(\lambda) = \argmin_{\substack{\beta}} {[} RSS(\beta) + J(\beta; \lambda) {]},
\end{equation*}
where $\lambda$ is the regularization coefficient and is usually
selected empirically (e.g. by cross validation).

We use ridge regression \cite{Hoerl1970-ux}, which imposes a penalty
on the $L_2$ norm of $\beta$, as in:
\begin{equation*}
J(\beta; \lambda) = \lambda \sum_i \beta_i^2
\end{equation*}
This penalty constrains the coefficients such that irrelevant ones
shrink towards zero without being eliminated from the model altogether.

Since the predictions of a linear model trained in such a way are
real values, not integers, the predictions have to be mapped back
onto the integer-valued score range.  One method for determining
the integer-valued score for a real-valued prediction $\hat{y}$
that lies between adjacent scores $z_1$ and $z_2$ is to apply
thresholds as follows:
\begin{equation}
\label{eq:threshold}
  Score(\hat{y}, z_1, z_2) = \begin{cases}
    z_1 & \text{if $\hat{y} \leq \frac{z_1 + z_2}{2}$;} \\
    z_2 & \text{otherwise}.
  \end{cases}
\end{equation}
Other reasonable approaches to the problem are to use ordinal
logistic regression or to choose the thresholds based on performance
on a validation set.  Here we use Equation \ref{eq:threshold}.

\section{Optimal design algorithms}

In our experiments, we start with a set of $n$ candidate essays.
We wish to sample a subset of $m$ of these essays that will give
us a reasonably accurate predictive model after a human scores them.

More formally, let $\mathbf{X} \in \mathbb{R}^{n \times p}$ be the
$n$ length-$p$ feature vectors of the essays.  Call this the
\textit{feature space}.  The $i$th row of $\mathbf{X}$ is $x_i$ and
the $j$th column is $x_j^T$.  Assume we lack the \textit{target
variable} $\mathbf{Y} \in \mathbb{R}^{n \times 1}$.  Also let $\Xi$
be the set of all subsets of rows of $\mathbf{X}$; this is the
\textit{design space}.  A \textit{design} $\xi$ is an $m$-row member
of $\Xi$.  The goal of the algorithms described in this section is
to find $\xi$ such that, when it is agumented with $\mathbf{Y}_\xi$
-- the $m \times 1$ matrix containing the target for the feature
vectors in $\xi$ -- a supervised model that learns $P(\mathbf{Y_\xi}
| \mathbf{\xi})$ has better predictive performance on unseen feature
vectors than one would expect if $\xi$ were chosen at random.

\begin{figure}[!htbp]
\centering
\includegraphics[height=1\textwidth]{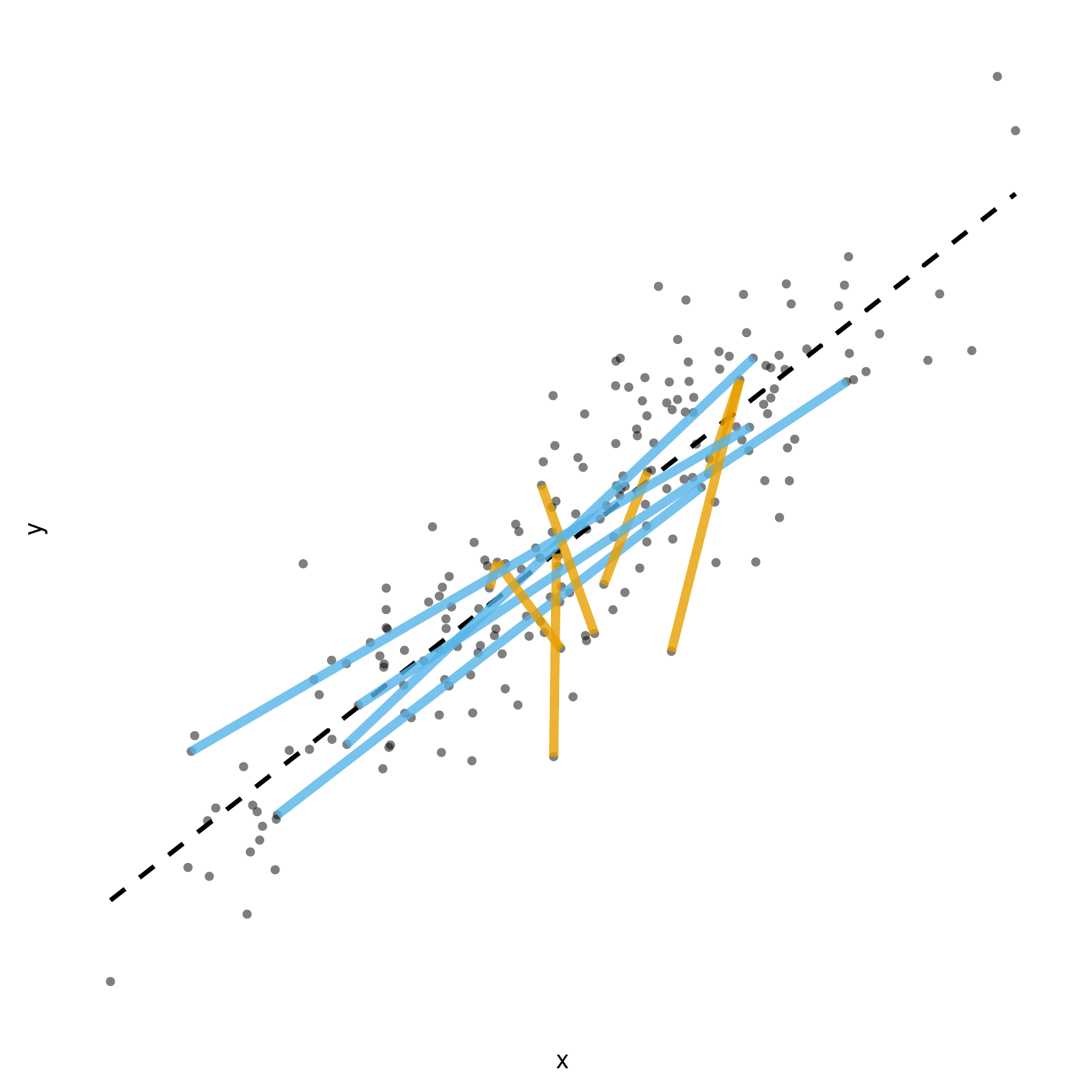}
\caption{
When fitting a regression line on a subset of the data, using distant points increases the probability that the fit will be close to the best fit.  The regression line for all of the data is the dashed line.  For contrast, lines connecting near and distant points are shown in different colors.  The data are two hundred simulated points, with $x$ drawn from a random normal distribution and $y = 1.5x + \epsilon$.}
\label{fig:regression}
\end{figure}

In our case the model $P(\mathbf{Y_\xi} | \mathbf{\xi})$ is linear
regression.  Consider a regression with feature $x$ and target $y$.
If we only allow an algorithm to choose two (in this case,
unidimensional) feature vectors, which choices might result in a
better regression model?  The consequences of different choices are
illustrated in Figure \ref{fig:regression}, where one can see that
the greater the distance between the $x$ values, the greater the
probability that the model will perform almost as well as one trained
with all of the data.  This is the basic concept behind effective
sampling of feature vectors for regression models.  It motivates
our selection of the algorithms we discuss in this section.  The choices
they make constitute a spectrum from choosing maximally distant
feature vectors to choosing uniformly distant ones.

\begin{figure*}[!htbp]
\centering
\subfloat[Fedorov with $D$-optimality]{
  \includegraphics[height=0.35\textwidth]{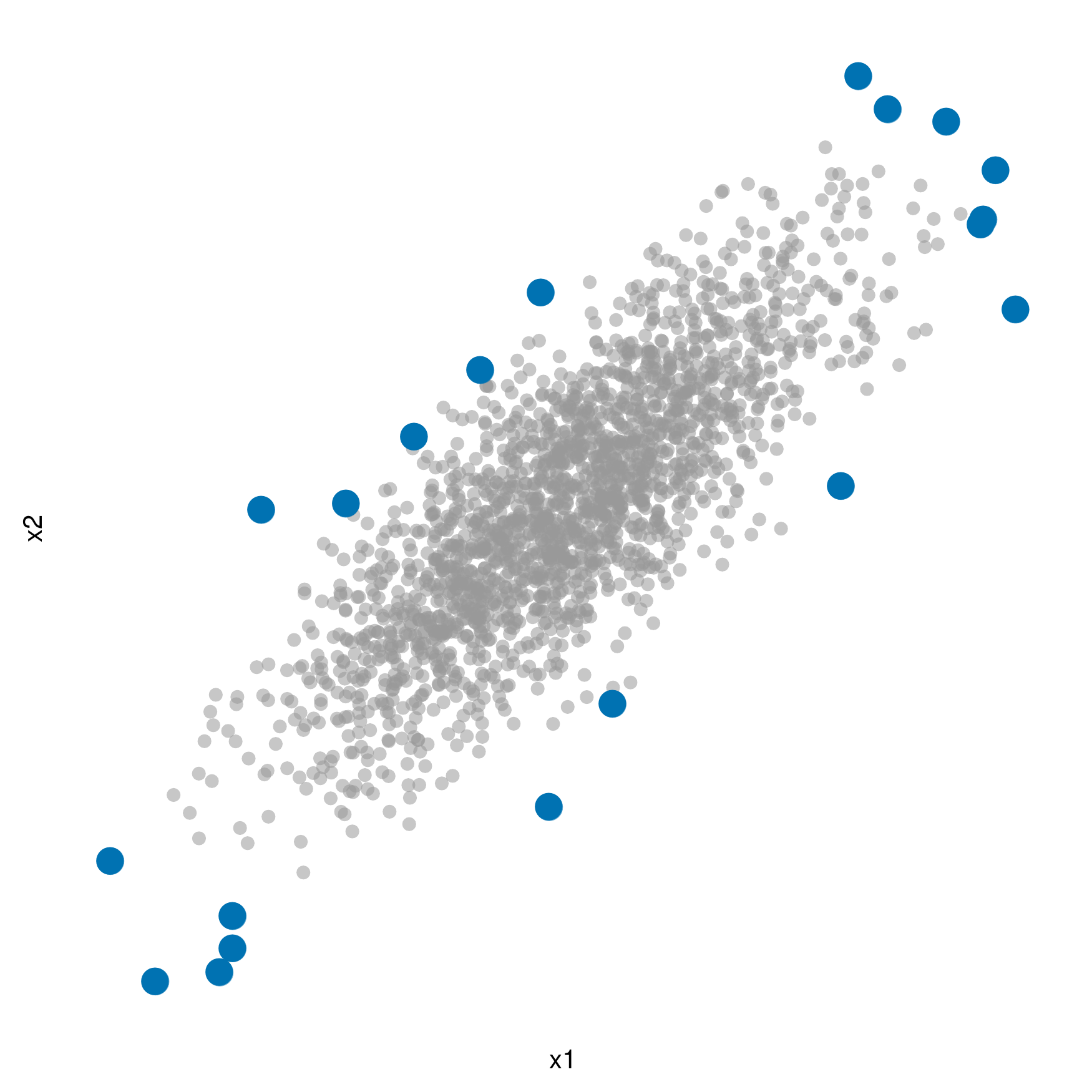}
  \label{fig:fedorov}
} 
\subfloat[Kennard-Stone]{
  \includegraphics[height=0.35\textwidth]{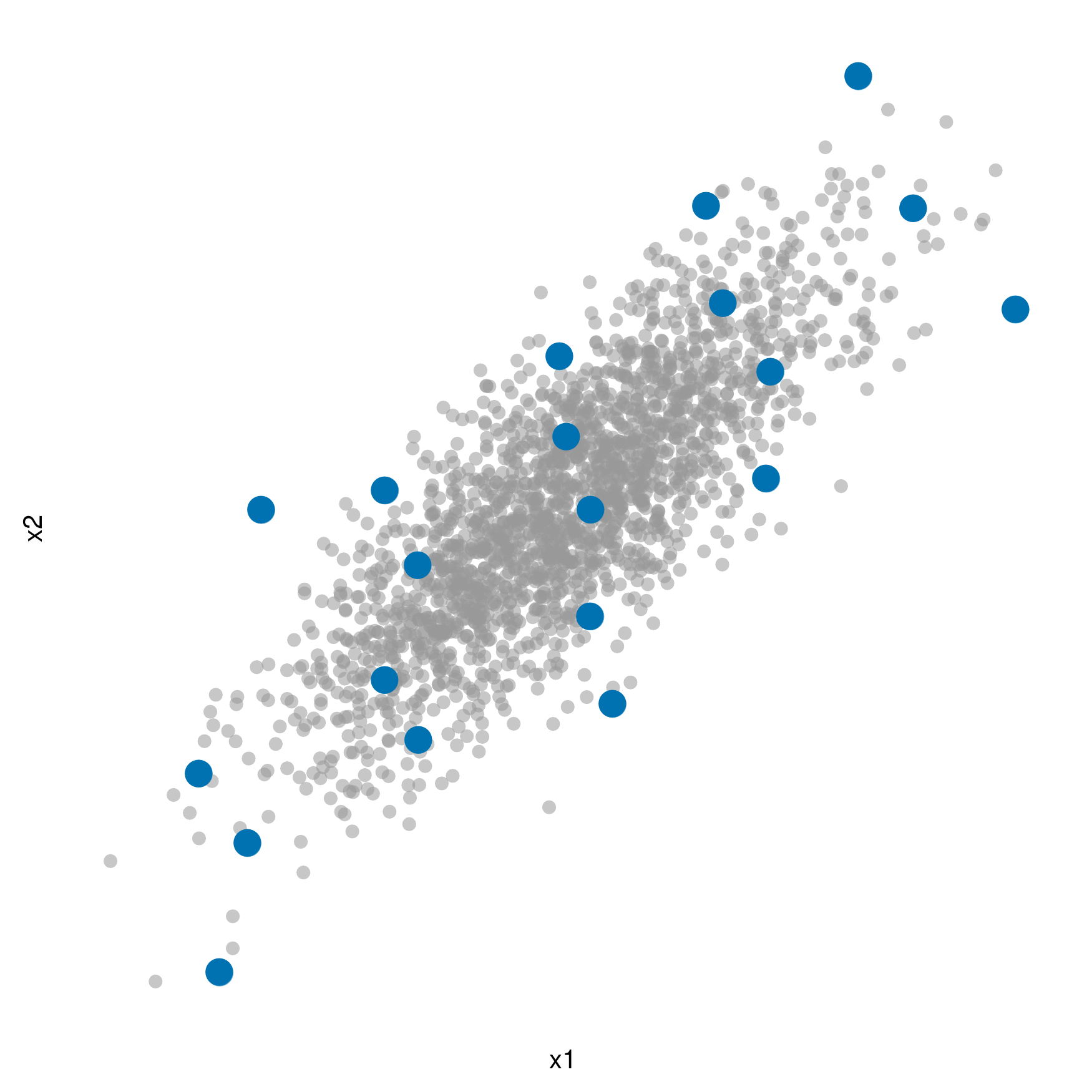} 
  \label{fig:kennard-stone}
} 
\subfloat[$K$-means]{
  \includegraphics[height=0.35\textwidth]{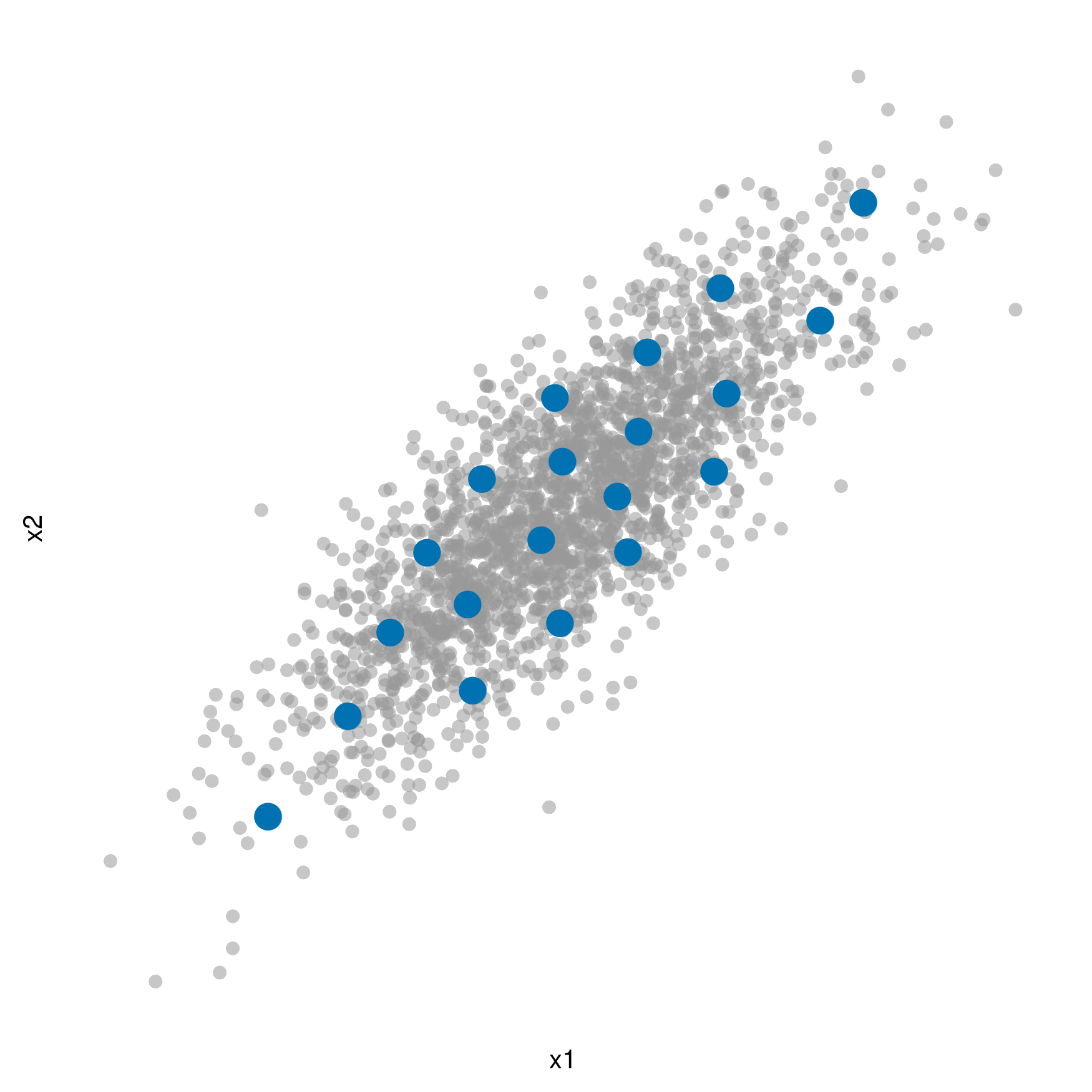} 
  \label{fig:k-means}
}
\caption{The points chosen by algorithms are distributed differently in the feature space.  Here the larger points are twenty points selected by an algorithm.  The data are simulated.  In each figure, a point is one out of a thousand samples drawn at random from a two-dimensional multivariate normal distribution.}
\label{fig:choices}
\end{figure*}

The R statistical language \cite{R} has implementations of the
algorithms we discuss in this section.  The implementation
of the Fedorov algorithm we used is in the AlgDesign package \cite{AlgDesign}.
The prospectr package \cite{prospectr} contains implementations of
Kennard-Stone and $k$-means sampling.

\subsection{Fedorov exchange algorithm (with $D$-optimality)}

The Fedorov exchange algorithm \cite{Fedorov1972-pg} is a greedy
stepwise algorithm for finding $\xi$.  Its purpose is to optimize
various optimal design criteria -- $D$-optimality, $A$-optimality,
$I$-optimality, or others \cite{John1975-ty}.  Most criteria are
some function of the information matrix of $\xi$, defined as $M(\xi)
= \frac{1}{m}\xi^T \xi$.  $D$-optimality, for instance, searches
for the $\xi$ that maximizes the determinant $det~M(\xi)$.  In
experiments with internal data sets, we didn't see major differences
among $D$-optimality, $A$-optimality, or $I$-optimality, but the
$D$-optimality criterion did tend to perform somewhat better, so for
simplicity of exposition we limit ourselves to $D$-optimality in
the experiments we conduct with the ASAP essay scoring data.

The first step of the Fedorov algorithm is to randomly initialize
$\xi$ with $m$ rows from $X$.  Subsequent steps replace a row in
$\xi$ with a row from $X$ that is not already in $\xi$.  The swapped
rows are chosen greedily to optimize the design criterion.  The
algorithm can get stuck in a local optimum, depending on the initial
choice of $\xi$.  The remedy for this is to run the algorithm several
times with different initial states and to choose the best one.
The resulting design $\xi$ consists of feature vectors that are
maximally distant from the centroid of the feature space, as can
be seen in Figure \ref{fig:fedorov}.

\subsection{Kennard-Stone algorithm}

The initialization step of the Kennard-Stone algorithm \cite{Kennard1969-ac} is
similar to D-optimality in that it chooses two feature vectors at the
periphery of the feature space.  Thereafter, however, it chooses
feature vectors so as to distribute them uniformly.  More precisely, let
$d(u, v)$ be the distance between $u$ and $v$.  The first step of
the Kennard-Stone algorithm is to choose indices $i, j$ such that
\begin{equation*}
\argmax_{i,j \in 1 \dots n} d(x_i, x_j),
\end{equation*}
then to initialize $\xi$ with $x_i, x_j$ and set $I_\xi = \{i,j\}$.  Define
\begin{equation*}
\Delta_{x_j}(\xi) = \underset{i \in I_\xi}{\min}~d(x_j, \xi_i).
\end{equation*}
On each subsequent step the index $j$ is added to $I_\xi$ and $x_j$
is added to $\xi$:
\begin{equation*}
\argmax_{j \notin I_\xi} \Delta_{x_j}(\xi).
\end{equation*}
After initialization, the point that is furthest from the point 
in the existing design are added at each step.  The distances can be computed in
any metric space.  Here they are computed in Mahalanobis
space.  Figure \ref{fig:kennard-stone} shows that a few of the
feature vectors n the design are maximially distant from the centroid
of the feature space and the rest are maximally distant from one
another.

\subsection{$k$-means sampling algorithm}

This is simply the $k$-means clustering algorithm with an extra
step at the end: after determining the final centroids, the design
is determined by choosing the observation closest to each of the $k$
centroids.  Thus when searching for a design of size $m$, we set $k=m$.

Here the design is approximately uniformly distributed throughout
the feature space in Figure \ref{fig:k-means}.  This design does
not include feature vectors at the periphery of the feature space,
because the centroid of a cluster to which a peripheral point belongs
will almost always be closer to some other point that is not
peripheral.  The uniformity on display here is in part an artifact
of the simulated data.   In real data the feature vectors would be
distributed less than smoothly, so the centroids of the final clusters
would be distributed less than uniformly.

\subsection{Persistence of selections}

\begin{figure}[!htbp]
\centering
\includegraphics[width=1\textwidth]{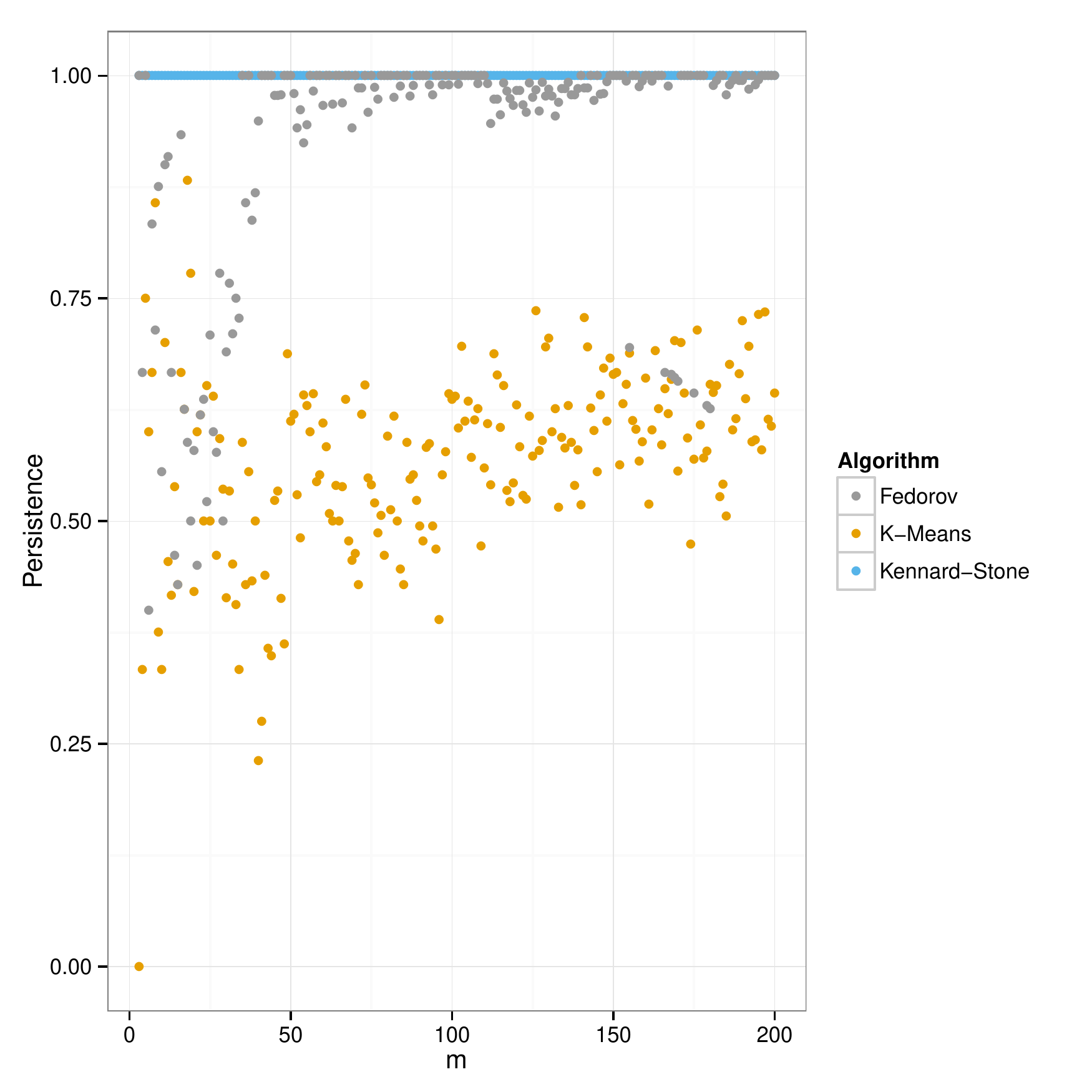}
\caption{The persistence of selections from ASAP AES essay set 1.  The Kennard-Stone algorithm's persistence is 1 everywhere.}
\label{fig:persistence}
\end{figure}

An important question is the degree to which the feature vectors selected
by these algorithms persist across sequential values of $m$.  If
the design $\xi$ at $m-1$ is always included in $\xi$ at $m$, its
selections persist perfectly.  The degree of persistence can be
measured as 
\begin{equation}
Persistence = \frac{|\xi_m \cap \xi_{m+1}|}{|m|}.
\end{equation}
The persistence of selections for increasing values of $m$ is shown
in Figure \ref{fig:persistence}.  Kennard-Stone starts with the
same two maximally distant feature vectors and adds further ones sequentially
according to a deterministic procedure, so its selections are
perfectly persistent.  The Fedorov algorithm runs several times
with different starting points in order to avoid getting stuck in
local optima.  This results in some loss of persistence, which
appears to improve as $m$ increases.  In the range $150 < m < 175$,
however, the Fedorov algorithm's persistence is erratic, suggesting
the existence of multiple local optima that are nearly indistinguishable
according to the optimality criterion.  The $k$-means algorithm
exhibits the lowest persistence of the three algorithms.  This is
expected, as the final centroids will change as $k$, the number of
clusters, increases.

\section{Evaluation}

To evaluate the performance of the algorithms, the ideal would be
to have access both to an effectively infinite number of unscored
essays and to an oracle that could provide scores upon request.
Since we have finite data, we simulate a much larger amount of data
by repeatedly sampling half of the entire training set (without
replacement) before choosing feature vectors to be scored by humans.
Of course, since we already have the scores, the human-scoring step
of our evaluation doesn't actually occur.  The algorithms don't
look at the scores, however, so no information leaks unfairly into
the process.

We let the desired training set size $m$ range from 10 to 100 in
increments of 10.  This allows us to simulate an operational
environment where new essays arrive over time.  After fixing $m$
and sampling half of the entire training set, we use each
algorithm to choose a subset of $m$ feature vectors, pretend to have
the corresponding essays scored by humans, and train one regression
model with each algorithm's chosen subset.  A model's performance
on the test set is the evaluation criterion -- specifically, the
Pearson correlation coefficient of its rounded predictions with the
scores of the human raters.  Since test set essays are the same for
all models, the performance of each model is comparable regardless
of the size of the training set.  To provide smoothed estimates,
we averaged the correlations for each of the 300 models trained
using a particular combination of training set, sampling algorithm,
and size $m$.

The baseline design algorithm is random sampling without
replacement.  For a given design size $m$, the baseline algorithm
selects a design $\xi_m$ of unique feature vectors from $X$ at random.
Models trained using a design chosen by an optimal design algorithm
are expected to perform better than ones trained using this
baseline.  Assuming an algorithm performs better than the baseline,
the key question is how its performance over the baseline changes
as $m$ increases.  The hallmark of an effective optimal design
algorithm is that it prefers to choose the essays that are most
informative to the model; this should manifest itself as a
\textit{larger} margin above the baseline when $m$ is small.  As
$m \rightarrow n$ (where $n$ is the number of feature vectors from
which to choose), the margin should go to 0.

\section{Results}
Table \ref{tab:percent-change} shows the percent change of the
average of each algorithm's models over the baseline.  To test
for significance, we transform the correlations using the Fisher
Z-transform before applying the two-sample t-test at the 0.001
level.  The test is an approximation, as the normality assumption
of the t-test does not hold, but the correlation coefficients are
unimodal and the sample sizes are relatively large, so we assume
that the approximation is quite good.  For interpretability, the
percent change is calculated using the Pearson correlation.  The
means of Fedorov and Kennard-Stone results are significantly different
from the baseline for most ASAP AES essay sets and training set
sizes Fedorov and Kennard-Stone, with the exception of the anomalous
essay set 4.  The $k$-means results are more often indistinguishable
from the baseline.

\begin{figure}[!htbp]
\centering
\includegraphics[width=1\textwidth]{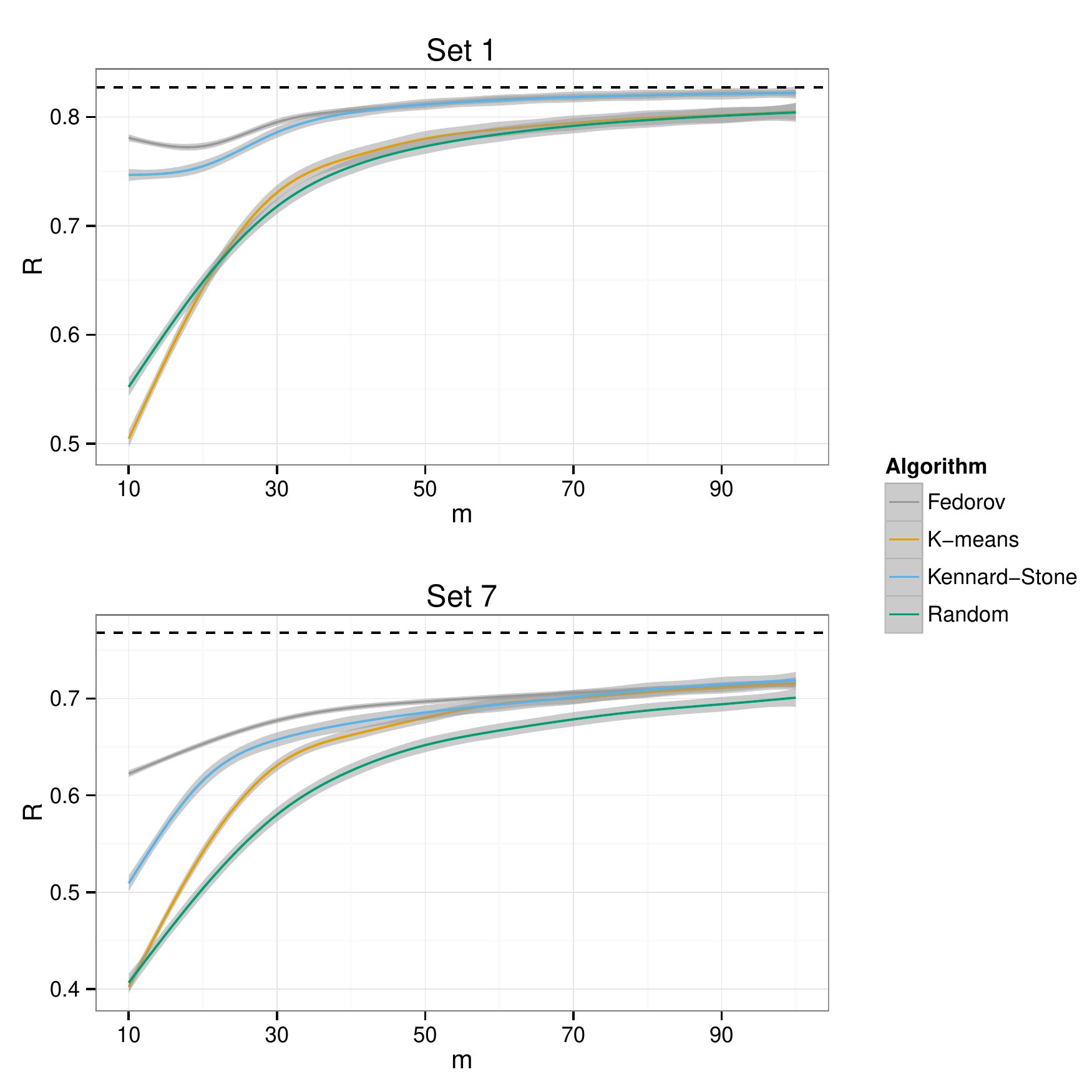}
\caption{The performance of each algorithm on two test sets.
Lines are the smoothed average of 300 models with 95\%
confidence intervals of the mean.  
The X axis, $m$, is the number of essays selected.  The Y axis, $R$,
is the Pearson correlation between the model and human raters 
on the test set.  
The dashed horizontal line is the mean test
set performance when training with all of the data available
\textit{in an iteration}.}
\label{fig:ridge-bootstrap}
\end{figure}

To achieve parity with the baseline at $m=100$, the Fedorov algorithm
typically needs to select only about 50 essays.  With essay set 1,
for instance, it only needs to select 10 essays, as can be seen in
Figure \ref{fig:ridge-bootstrap}.

\begin{table}[!htbp]
\centering
\subfloat[Fedorov]{
  \label{tab:percent-change-fedorov}
  \centering
  \begin{tabular}{lcccccccccc}
  \hline
  Set & \multicolumn{10}{c}{$m$} \\
  \hline
      & 10 & 20 & 30 & 40 & 50 & 60 & 70 & 80 & 90 & 100 \\ 
  \hline
   1  & 42  &  18  &   10  &   7 &   5 &   4 &   3 &   2 &   2 &   2 \\ 
   2a & 86  &  69  &   47  &  37 &  26 &  26 &  22 &  18 &  18 &  15 \\ 
   2b & 94  &  68  &   48  &  39 &  32 &  27 &  26 &  23 &  18 &  17 \\ 
   3 &  26  &  18  &   16  &  10 &   8 &   8 &   5 &   5 &   4 &   4 \\ 
   4 &  25  &  7  &    \textbf{\underline{4}}  & \textbf{\underline{3}} &   5 &   4 &   \textbf{\underline{3}} & \textbf{\underline{3}} & \textbf{\underline{4}} & \textbf{\underline{2}} \\ 
   5 &  25  &  12  &   7  &   4 &   3 &   2 &   1 &   1 &   1 &   1 \\ 
   6 &  59  &  30  &   17  &  12 &   8 &   7 &   5 &   4 &   3 &   3 \\ 
   7 &  54  &  29  &   16  &  10 &   6 &   5 &   4 &   2 &   2 &   2 \\ 
   8 & 106  &  84  &   62  &  93 &  80 &  65 &  74 &  84 &  50 &  63 \\ 
  \hline
  \end{tabular}
}

\subfloat[Kennard-Stone]{
  \label{tab:percent-change-kennard-stone}
  \centering
  \begin{tabular}{lcccccccccc}
  \hline
  Set & \multicolumn{10}{c}{$m$} \\
  \hline
      & 10 & 20 & 30 & 40 & 50 & 60 & 70 & 80 & 90 & 100 \\ 
  \hline
    1 &  36 &  15 &   9 &   6 &   5 &   4 &   3 &   2 &   2 &   2 \\ 
    2a &  62 &  70 &  48 &  36 &  26 &  26 &  21 &  18 &  18 &  15 \\ 
    2b &  70 &  67 &  45 &  36 &  30 &  25 &  25 &  22 &  17 &  17 \\ 
    3 &  25 &  13 &  13 &   9 &   7 &   9 &   6 &   5 &   5 &   4 \\ 
    4 & -33 &  \textbf{\underline{-2}} &  \textbf{\underline{-1}} &   \textbf{\underline{0}} &   \textbf{\underline{1}} &   \textbf{\underline{2}} &   \textbf{\underline{1}} &   \textbf{\underline{2}} &   \textbf{\underline{2}} &   \textbf{\underline{1}} \\ 
    5 &  \textbf{\underline{-5}} &  11 &   7 &   5 &   4 &   3 &   2 &   1 &   1 &   1 \\ 
    6 &  52 &  28 &  17 &  11 &   8 &   6 &   4 &   4 &   2 &   3 \\ 
    7 &  25 &  22 &  11 &   8 &   5 &   4 &   3 &   2 &   3 &   2 \\ 
    8 &  52 &  66 &  45 &  54 &  73 &  53 &  69 &  85 &  53 &  62 \\ 
  \hline
  \end{tabular}
}

\subfloat[$K$-means]{
  \label{tab:percent-change-k-means}
  \centering
  \begin{tabular}{lcccccccccc}
  Set & \multicolumn{10}{c}{$m$} \\
  \hline
      & 10 & 20 & 30 & 40 & 50 & 60 & 70 & 80 & 90 & 100 \\ 
  \hline
    1 &  -8 &  \textbf{\underline{-1}} &   \textbf{\underline{1}} &   \textbf{\underline{0}} &   \textbf{\underline{1}} &   \textbf{\underline{0}} &   \textbf{\underline{0}} &   \textbf{\underline{0}} &   \textbf{\underline{0}} &   \textbf{\underline{0}} \\ 
    2a &  30 &  42 &  32 &  23 &  17 &  19 &  16 &  13 &  14 &  12 \\ 
    2b &  53 &  40 &  32 &  28 &  23 &  19 &  20 &  18 &  14 &  14 \\ 
    3 &  \textbf{\underline{-4}} &   \textbf{\underline{0}} &   5 &   \textbf{\underline{3}} &   \textbf{\underline{1}} &   3 &   1 &   1 &   1 &   1 \\ 
    4 &  13 &  14 &  13 &   7 &   8 &   6 &   4 &   4 &   \textbf{\underline{3}} &   \textbf{\underline{0}} \\ 
    5 &   \textbf{\underline{4}} &   9 &   5 &   4 &   3 &   2 &   1 &   1 &   1 &   1 \\ 
    6 &  15 &   8 &  10 &   9 &   7 &   6 &   5 &   4 &   3 &   3 \\ 
    7 &   \textbf{\underline{0}} &   7 &   8 &   6 &   4 &   4 &   3 &   2 &   2 &   2 \\ 
    8 &  46 &  52 &  45 &  41 &  40 &  33 &  40 &  43 &  \textbf{\underline{22}} &  25 \\ 
  \hline
  \end{tabular}
}
\caption{Percent change over baseline of the mean performance of models trained with essays selected by optimal design algorithms.  Statistically insignificant differences are shown in bold.  Significance is determined by a two-sample t-test of means of $z$-transformed correlations with a 0.001 significance level.  
}
\label{tab:percent-change}
\end{table}

\begin{figure}[!htbp]
\centering
\includegraphics[width=1\textwidth]{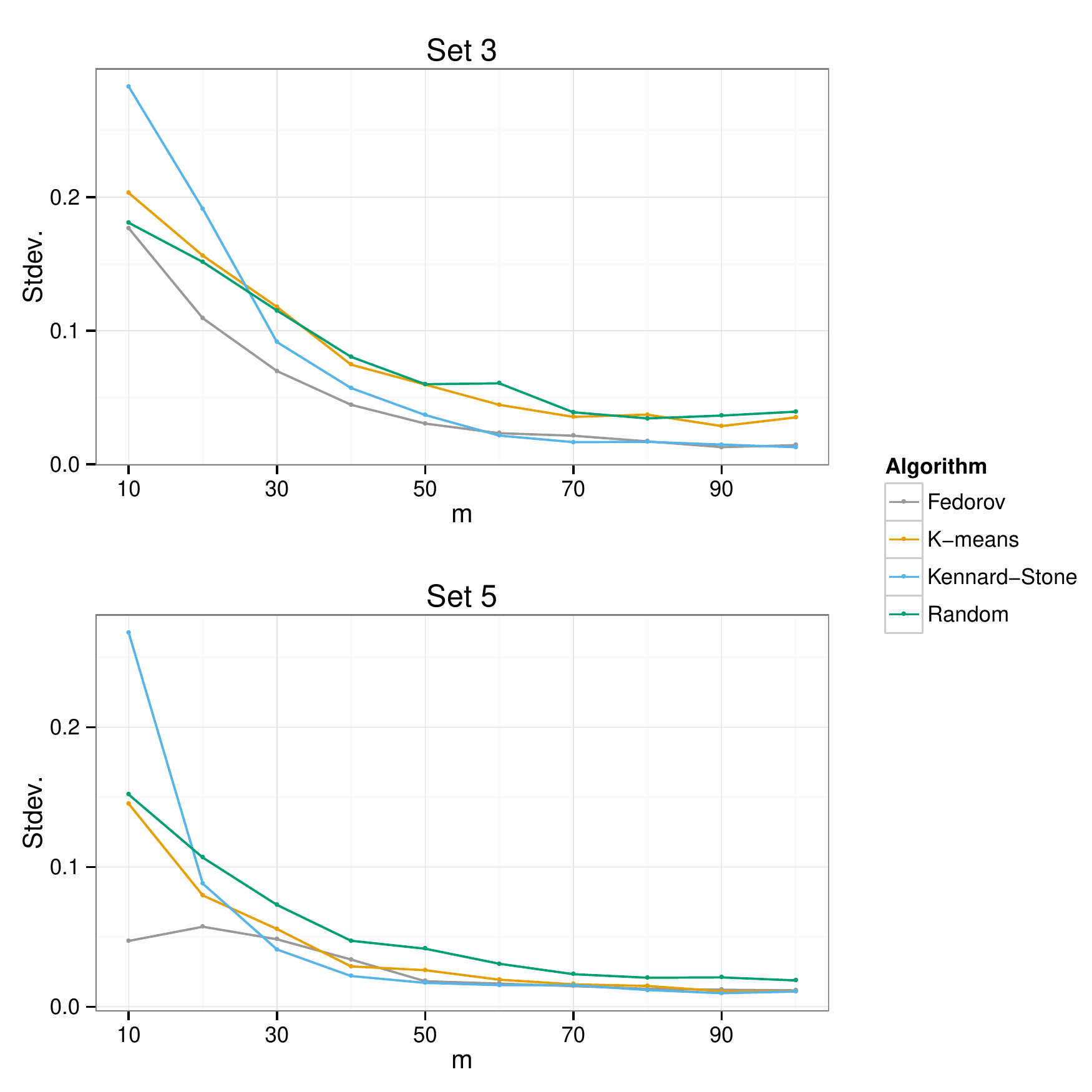}
\caption{Standard deviations of each algorithm's performance on two test sets.  The Y axis is the standard deviation of the Pearson correlation coefficients of 300 models trained with essays selected by an algorithm.  The X axis, $m$, is the number of essays selected.}
\label{fig:ridge-bootstrap-stdev}
\end{figure}

With respect to average performance, the Kennard-Stone algorithm
tends to perform well on the same sets as the Fedorov algorithm.
The confidence intervals in Figure \ref{fig:ridge-bootstrap} aren't
prominent enough to illustrate a notable difference.  The standard
deviations of the Kennard-Stone models tend to be greater, and for
some test sets are even worse than the baseline, as can be seen in
Figure \ref{fig:ridge-bootstrap-stdev}.  This suggests that the
Fedorov algorithm is the most robust: models trained with the essays
it selects have the best average performance and the least variance.
It also supports the notion -- illustrated in for 1-d data in Figure
\ref{fig:regression} -- that the periphery of the feature space is
the most effective region from which to choose feature vectors.

We should expect the Fedorov algorithm to perform best when the
underlying process that generates the data is best fit using a
linear model.  Overall, the ranking of optimal design algorithms
-- from best to worst -- conforms to this expectation: Fedorov,
Kennard-Stone, and $k$-means.  The only anomaly is essay set 4,
where the $k$-means models perform best for $10 < m < 60$, with
respect both to the mean of the Pearson correlation coefficient and
its standard deviation.  Two possible explanations of this are noise
introduced by human raters or an insufficient set of features in
the feature space.

Based on the relative performances of the three optimal design
algorithms, we strongly favor the use of the Fedorov algorithm
with $D$ optimality.  The Kennard-Stone algorithm's performance is
a close second, but the models trained with the essays it selects
tend to have greater variance and when $m$ is small it doesn't
approach the impressive performance of the Fedorov algorithm.

\section{Conclusion}
Effective sampling disciplines the unruly process of obtaining
training sets for AWE models.  Our simulations show that models
trained using a relatively small number of intelligently-chosen
essays tend to perform well on out-of-sample essays, at times almost
as well as a model trained on an entire training set.  In several
of the data sets, training with 30-50 essays yields approximately
the same performance as a model trained with hundreds of essays.
The key implication is that it is possible to minimize the number
of human scores necessary for training an AWE model without unduly
sacrificing accuracy.

These results also have implications for designing systems that allow
rapid integration of new AWE scoring prompts into large-scale
systems.  One such approach is to train models incrementally.
For example, in a large MOOC, a professor may post a new writing
prompt for the students.  After the first few hundred students
submit their essay, the optimal design algorithm selects a subset
for the professor or teaching assistant to score.  Using this initial
training set, a scoring model can quickly be built that will start
automatically scoring student essays, so that students can receive
feedback on their writing.  As more essays are received from students,
the optimal design algorithm can select additional
essays for human scoring that will be most informative to help
refine the accuracy of the scoring model. The model can then be
automatically retrained with even greater accuracy of scoring.

When a model is retrained as scores arrive into a data collection
system, using optimal design to select the essays from a large pool
will ensure that the initial training set consists only of essays
that are likely to result in a reasonable model.  It will also both
ensure that subsequently-added essays are maximally informative and
minimize the duplication of the effort of human raters.

Optimal design is necessary but not sufficient for making incremental
training of AWE models effective and accurate.  While it minimizes
the sampling noise in a training set, it doesn't guarantee a minimum
level of performance.  To ensure that models are not deployed
prematurely, a validation set of essays should first be selected
from the pool and scored by humans.  Measuring model performance
with a validation set will ensure that the model performs well
enough before it is deployed.  Each time a new human score enters
the data collection system, the model can be retrained, and its
performance on the validation set can be computed.  If the model
doesn't perform well enough to be deployed to a production environment,
additional essays can be selected to be scored.  This process should
repeat until the model reaches an acceptable level of performance.

In experiments with internal data sets, we noticed that when a model
trained on the entire training set performed to our satisfaction,
optimal design performed well too.  The conditions that caused a
model trained on the entire training set to perform poorly also
caused the performance of optimal design to degrade.  The most
common cause of not meeting the conditions for good model performance
is noise due either to poor agreement among human raters or model
misspecification.  An example of model misspecification is when the
feature set does not capture textual semantics and the rubric
instructs the human raters to give a higher score to an essay if
it covers some set of topics\footnote{There has been some recent
work on active learning for misspecified models
\cite{Sugiyama2005-jr,Bach2007-ub}, but their thrust has been
primarily theoretical and do not appear to be immediately helpful
for this problem.}.  Our recommended solution to avoiding model
misspecification, which we don't evaluate here, is to be generous 
with one's feature set and to let a regularizer shrink uninformative
features.  We expect the results we report here to generalize to
other tasks and populations as long as the conditions are right for
a model trained on the entire training set to perform satisfactorily.

While these results are promising for the scoring of essays, we
expect that it will be less effective with short answer data sets.
Short answer data sets are often modeled using a learning algorithm
that employs a set of decision trees, such as random forest.  The
feature vectors are often different, too, because such models perform
well when they know exactly which words a student used.  Consequently,
it is common for the feature vectors to be rows of a \textit{document-term
matrix} -- an occurrence matrix with terms as columns and documents
as rows -- constructed from the training set responses.  In that
representation, the feature vectors are sparse and relatively long.
Optimal design and active learning work best when the assumptions
of the sampling algorithm agree with the assumptions of the subsequent
supervised learning algorithm.  Thus, using the Fedorov algorithm
to select a subset of short responses to be scored by humans when
the supervised model is random forest trained with a document-term
matrix is not guaranteed to yield the best results.

The use of optimal design for AWE training sets can be seen as a
form of manufacturing process control in which the amount of
information provided by a human score is maximized and its variance
minimized.  From this perspective, the goal of our use of optimal
design is primarily to make the output of the human scoring process
-- that is, the quantity of information in a single human score --
more predictable.  The effect of this is to enable safe and reliable
incremental training of AWE models.  Overall, the approach solves
a barrier to the adoption of AWE into large-scale formative and
summative systems, allowing the computer to minimize the amount of
human effort needed to collect and score essays by choosing effectively
which essays humans need to score.


\bibliographystyle{alpha}
\bibliography{arxiv}

\begin{thebibliography}{{R C}14}

\bibitem[AB06]{Attali2006-nz}
Yigal Attali and Jill Burstein.
\newblock Automated essay scoring with e-rater® v. 2.
\newblock {\em The Journal of Technology, Learning and Assessment}, 4(3), 2006.

\bibitem[Bac07]{Bach2007-ub}
F~R Bach.
\newblock Active learning for misspeciﬁed generalized linear models.
\newblock {\em Adv. Neural Inf. Process. Syst.}, 2007.

\bibitem[edXa]{edXdiscern}
\url{https://github.com/edx/discern}.
\newblock Accessed: 2013-10-7.

\bibitem[edXb]{edXdiscerngithub}
\url{https://github.com/edx/discern/blob/master/ml_grading/ml_model_creation.py}.
\newblock Accessed: 2013-10-7.

\bibitem[edXc]{edXease}
\url{https://github.com/edx/ease}.
\newblock Accessed: 2013-10-7.

\bibitem[edXd]{edXorg}
\url{http://edx.org}.
\newblock Accessed: 2013-10-7.

\bibitem[ETS]{ETSCriterion}
\url{https://criterion.ets.org}.
\newblock Accessed: 2013-10-13.

\bibitem[Fed72]{Fedorov1972-pg}
V~V Fedorov.
\newblock {\em Theory Of Optimal Experiments}.
\newblock Probability and mathematical statistics. Elsevier Science, 1972.

\bibitem[FSLL13]{Foltz2013-li}
Peter~W Foltz, Lynn Streeter, Karen~E Lochbaum, and Thomas~K Landauer.
\newblock Implementation and applications of the intelligent essay assessor.
\newblock In Mark~D Shermis and Jill Burstein, editors, {\em Handbook of
  Automated Essay Evaluation: Current Applications and New Directions}, pages
  68--88. Taylor \& Francis, 2013.

\bibitem[HK70]{Hoerl1970-ux}
Arthur~E Hoerl and Robert~W Kennard.
\newblock Ridge regression: Biased estimation for nonorthogonal problems.
\newblock {\em Technometrics}, 12(1):55--67, 1~February 1970.

\bibitem[JD75]{John1975-ty}
R~C~St John and N~R Draper.
\newblock {D-Optimality} for regression designs: A review.
\newblock {\em Technometrics}, 17(1):15--23, 1975.

\bibitem[Kaga]{KaggleAES}
The hewlett foundation: Automated essay scoring | kaggle.
\newblock \url{http://www.kaggle.com/c/ASAP-AES}.
\newblock Accessed: 2013-10-7.

\bibitem[Kagb]{KaggleSAS}
The hewlett foundation: Short answer scoring | kaggle.
\newblock \url{http://www.kaggle.com/c/ASAP-SAS}.
\newblock Accessed: 2013-10-7.

\bibitem[KS69]{Kennard1969-ac}
R~W Kennard and L~A Stone.
\newblock Computer aided design of experiments.
\newblock {\em Technometrics}, 11(1):137--148, 1~February 1969.

\bibitem[LLF03]{Landauer2003-vm}
Thomas~K Landauer, Darrell Laham, and Peter~W Foltz.
\newblock Automatic essay assessment.
\newblock {\em Assessment in Education: Principles, Policy and Practice},
  10(3):295--308, 2003.

\bibitem[Mac92]{MacKay1992-gp}
D~MacKay.
\newblock {Information-Based} objective functions for active data selection.
\newblock {\em Neural Computation}, 4(4):590--604, July 1992.

\bibitem[PP68]{Page1968-ix}
EB~Page and DH~Paulus.
\newblock The analysis of essays by computer. final report.
\newblock 1968.

\bibitem[{R C}14]{R}
{R Core Team}.
\newblock {\em R: A Language and Environment for Statistical Computing}.
\newblock R Foundation for Statistical Computing, Vienna, Austria, 2014.

\bibitem[RGW06]{Rudner2006-qb}
Lawrence~M Rudner, Veronica Garcia, and Catherine Welch.
\newblock An evaluation of {IntelliMetric™} essay scoring system.
\newblock {\em The Journal of Technology, Learning and Assessment}, 4(4), 2006.

\bibitem[Set10]{Settles2010-wp}
B~Settles.
\newblock Active learning literature survey.
\newblock {\em University of Wisconsin, Madison}, 2010.

\bibitem[Smi18]{Smith1918-ho}
Kirstine Smith.
\newblock On the standard deviations of adjusted and interpolated values of an
  observed polynomial function and its constants and the guidance they give
  towards a proper choice of the distribution of observations.
\newblock {\em Biometrika}, 12(1/2):1--85, 1~November 1918.

\bibitem[SRL13]{prospectr}
Antoine Stevens and Leornardo Ramirez-Lopez.
\newblock {\em An introduction to the prospectr package}, 2013.
\newblock R package version 0.1.3.

\bibitem[Sug05]{Sugiyama2005-jr}
M~Sugiyama.
\newblock Active learning for misspecified models.
\newblock {\em Adv. Neural Inf. Process. Syst.}, 2005.

\bibitem[Whe14]{AlgDesign}
Bob Wheeler.
\newblock {\em AlgDesign: Algorithmic Experimental Design}, 2014.
\newblock R package version 1.1-7.2.

\bibitem[Wri]{WriteToLearn}
\url{http://writetolearn.net}.
\newblock Accessed: 2013-10-13.

\end{thebibliography}

\end{document}